%% file: main.tex
\documentclass[10pt,twocolumn,letterpaper]{article}

\usepackage{cvpr}              
\usepackage{algorithm}
\usepackage{algpseudocode}
\usepackage{stfloats}
\usepackage[accsupp]{axessibility}

\input{preamble}

\definecolor{cvprblue}{rgb}{0.21,0.49,0.74}
\usepackage[pagebackref,breaklinks,colorlinks,allcolors=cvprblue]{hyperref}

\def\paperID{11} 
\def\confName{CVPR}
\def\confYear{2026}

\title{Towards Design Compositing}

\author{
Abhinav Mahajan$^{1*,\dagger}$ \quad
Abhikhya Tripathy$^{1,\dagger}$ \quad
Sudeeksha Reddy Pala$^{2,\dagger}$ \quad
Vaibhav Methi$^{3,\dagger}$ \\
K J Joseph$^{4}$ \quad
Balaji Vasan Srinivasan$^{4}$ \\[4pt]
{\normalsize
$^1$Carnegie Mellon University \quad
$^2$IIT Kharagpur \quad
$^3$IIT Kanpur \quad
$^4$Adobe Research}
\\
{\tt\small \{abhinavm, abhikhyt\}@andrew.cmu.edu, \{josephkj, balsrini\}@adobe.com}
}

\begin{document}

\twocolumn[{%
\renewcommand\twocolumn[1][]{#1}%
\renewcommand{\thefootnote}{}
\maketitle
\begin{center}
\vspace{-15pt}
\includegraphics[height=0.27\textheight,keepaspectratio]{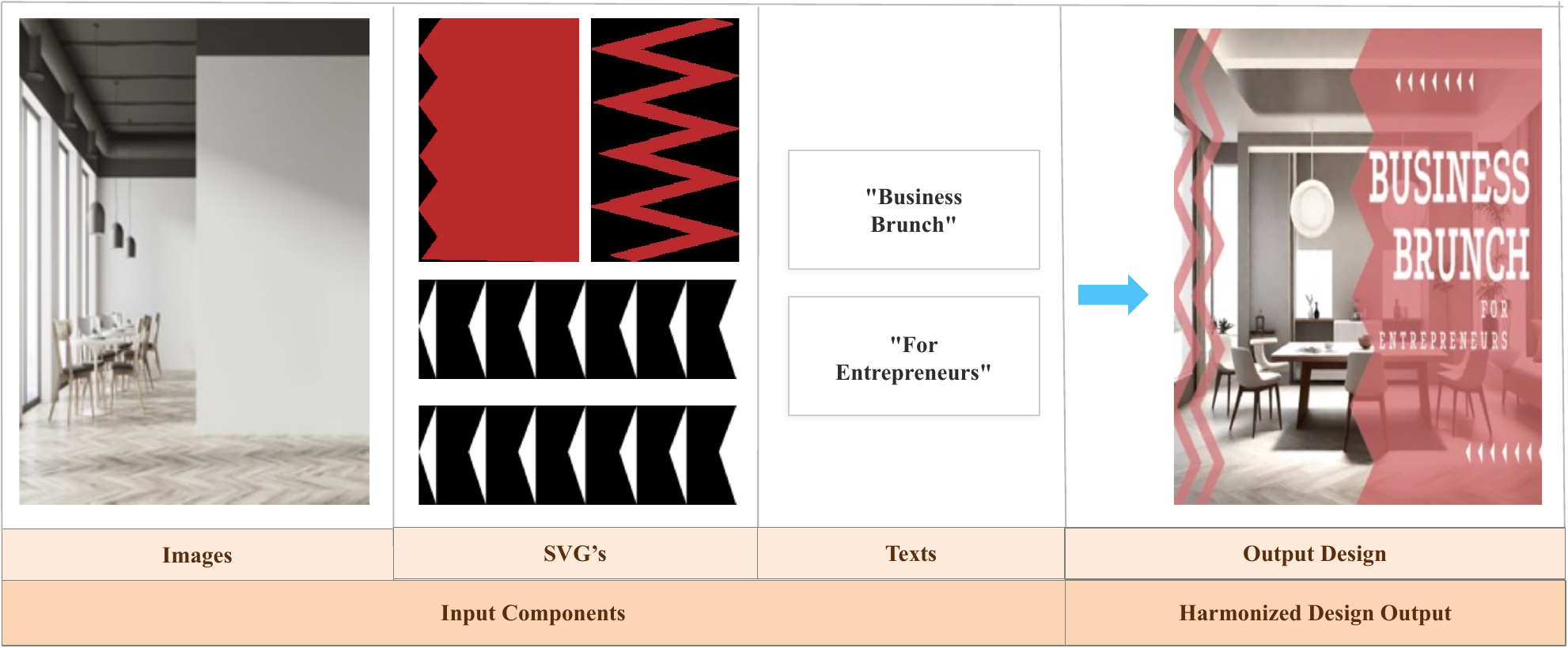}
\captionof{figure}{%
\textbf{Design Compositing with \methodname{}:}
We introduce \methodname{} (\textbf{G}rounded \textbf{I}dentity-preserving \textbf{S}tylized composi\textbf{T}ion), a compositing stage that transforms and harmonizes visual elements, images and SVGs, at their predicted
layout positions rather than naively pasting them, while preserving their semantic identity. Note how the image is stylized along with retaining the dining room scene and feels cohesive with the surrounding elements. The SVGs are similarly transformed for visual harmony. This is in contrast to existing methods which focus on layout prediction (spatial arrangement) and typography (font, color, and placement of text), without transforming the content itself.}
\label{fig:teaser}
\end{center}
}]

\begingroup
\renewcommand{\thefootnote}{\fnsymbol{footnote}}
\footnotetext[2]{Work done during internship/employment at Adobe Research.}
\endgroup

\begin{abstract}

Graphic design creation involves harmoniously assembling multimodal components such as images, text, logos, and other visual assets collected from diverse sources, into a visually-appealing and cohesive design. Recent methods have largely focused on layout prediction or complementary element generation, while retaining input elements exactly, implicitly assuming that provided components are already stylistically harmonious. In practice, inputs often come from disparate sources and exhibit visual mismatch, making this assumption limiting. We argue that identity-preserving stylization and compositing of input elements is a critical missing ingredient for truly harmonized components-to-design pipelines. To this end, we propose \textbf{\methodname{}}, a training-free, identity-preserving image compositor that sits between layout prediction and typography generation, and can be plugged into any existing components-to-design or design-refining pipeline without modification. We demonstrate this by integrating
\methodname{} with two substantially different existing methods, LaDeCo~\cite{lin2024elementsdesignlayeredapproach} and Design-o-meter~\cite{goyal2024design}. 
\methodname{} shows significant improvements in visual harmony and aesthetic  quality across both pipelines, as validated by LLaVA-OV and GPT-4V on aspect-wise ratings and pairwise preference over naive pasting. Project Page: \href{https://abhinav-mahajan10.github.io/GIST/}{\tt{abhinav-mahajan10.github.io/GIST/}}.

\end{abstract}

\section{Introduction}
\label{sec:intro}


Graphic designs are ubiquitous in daily life, appearing in advertisements, social media posts, posters, banners, presentations, and digital user interfaces. Two key factors determining the effectiveness of a graphic design are whether it is visually-appealing, eye-catching, coherent, and aesthetically cohesive, and second, whether it clearly communicates the designer's intended message. Designs that satisfy both criteria are generally more impactful and engaging.

However, creating such designs at scale is challenging since modern designers often work under tight deadlines and handle a large volume of design tasks. This makes it difficult to manually ensure both visual quality and consistency across outputs, which automated graphic design generation systems can help with by accelerating routine workflows while maintaining visual appeal and adherence to user intent. Such systems can also lower the barrier to entry for novice users by providing design capabilities that would otherwise require substantial artistic expertise.

An important instance of this is the components-to-design workflow. Graphic designers often assemble multimodal components such as images, text, logos, SVG elements, and other visual assets collected from diverse sources, into a cohesive aesthetic design. Recent works have made encouraging progress on automating workflow. Some methods focus on predicting the spatial arrangement of input elements \cite{inoue2023flexiblemultimodaldocumentmodels, inoue2023layoutdm, lin2024elementsdesignlayeredapproach}, while others generate complementary components that complete the design in harmony with the provided inputs \cite{zhang2025creatidesign}. 

A key characteristic of earlier works in this domain is that they preserve the input components exactly as provided. This constraint works well only under the assumption that the user-specified assets are already visually harmonious and need only be arranged appropriately or supplemented with missing or additional components. This assumption becomes a strong limitation in realistic settings, where input components are often collected from diverse sources and may differ substantially in color palette, rendering style, tone, texture, or overall visual appearance. In such cases, simply arranging elements that are not harmonious with respect to each other is insufficient to produce a harmonious design. The restriction of preserving inputs exactly contradicts the very objective of true design harmony.

We argue that true harmonized design generation from input components requires input-conditioned, identity-preserving component stylization and composition. Here, composition refers to adapting the appearance of existing design elements so that they become visually cohesive with the rest of the design while still preserving the semantic identity and designer intent associated with those elements. Thus, design composition may involve, for example, compositing image elements to better match a common style or color palette, or adjusting text typography attributes such as color and font to complement surrounding elements. Existing works have at best tackled stylization for textual elements, while largely overlooking the \textbf{composition of image-based elements}. This is a major research gap since image elements often dominate the overall aesthetics of a graphic design, therefore playing a critical role in perceived visual harmony and aesthetics.

To address this gap, we propose \textbf{\methodname{}: }Grounded Identity-preserving Stylized composiTion, a training-free, identity-preserving image composition method that sits between layout prediction and typography, and can be plugged into any existing components-to-design pipeline without modification. For complete frameworks like LaDeCo~\cite{lin2024elementsdesignlayeredapproach}, \methodname{} slots in as a compositing stage; for layout-only methods like Design-o-meter~\cite{goyal2024design} (which natively supports only repositioning texts, treating them as static renders), we can extend them toward full components-to-design generation by additionally incorporating typography prediction from works like COLE~\cite{jia2023cole} or FlexDM~\cite{inoue2023flexiblemultimodaldocumentmodels}, in both cases complementing strong existing models toward the larger goal of true design harmony. A key advantage is that it adapts an existing foundational MLLM without any additional finetuning, exploiting its architectural bottleneck to perform generative image composition with enhanced identity preservation.

\noindent Concretely, our key contributions are as follows:
\begin{itemize}
    \item A novel identity-preserving image composition method, addressing the most critical missing functionality for harmonized components-to-graphic design generation.
    \item We demonstrate the plug-and-play nature of this method by integrating it with strong existing layout and typography models within an end-to-end pipeline for automated harmonized design generation from components. We show this for two distinct pipeline setups.
    \item We provide quantitative and qualitative results that validate our claims and show the importance of image composition for visually cohesive design generation.
\end{itemize}

\section{Related Works}
\label{sec:related}

\textbf{Optimizing Specific Graphic Design Aspects:}
A well-researched line of work in refining graphic designs is layout generation, where design elements are spatially arranged for visual harmony and comprehensibility. Earlier works achieved this by maximizing energy functions \cite{o2015designscape, o2014learning} or using aesthetic scores \cite{bauerly2006computational, yang2016automatic, hurst2009review, jahanian2013recommendation, kong2022aesthetics++, patnaik2025aesthetiq, chen2024towards}, which suffered from limitations such as high computational costs and human annotation bias respectively. These methods also didn't scale well to complex and diverse designs. Some transformer-based approaches \cite{inoue2023flexiblemultimodaldocumentmodels, gupta2021layouttransformer, xie2021canvasemb, yamaguchi2021canvasvae} have been developed which improved the latency and diversity of layout generation. FlexDM \cite{inoue2023flexiblemultimodaldocumentmodels} employs multitask learning in a single transformer-based model to solve various design tasks, including layout and typography, but suffers when the number of masked tokens increases. Recently, layout generation methods have employed diffusion models \cite{chai2023two, hui2023unifying, inoue2023layoutdm, zheng2023layoutdiffusion, guerreiro2024layoutflow}, Large Language Models (LLMs) \cite{lin2024layoutprompter, tang2023layoutnuwa, seol2024posterllama, biswas2024docsynthv2}, and Large Multimodal Models (LMMs) \cite{zhu2024automatic, cheng2024graphic, kikuchi2024multimodal, lin2024elementsdesignlayeredapproach}. Although layout configuration is critical, our work goes a step further by addressing the fundamental challenge of composing and stylizing image-based elements at the predicted layout positions, which is imperative for true composition.

\noindent \textbf{Automated Graphic Design Generation:}
Recently, COLE \cite{jia2023cole} proposed a pipeline approach for generating designs from scratch given a user intention. They achieve this by training a design planner, which generates a structured JSON of the design contents, then render the visual assets using cascaded diffusion, and finally use a typography LMM for predicting the appropriate text data. OpenCOLE \cite{inoue2024opencole} is an open source implementation of COLE with some changes. Although these methods provide some edit-ability, designers cannot use input assets of their choice. IGD \cite{qu2025igd} formulates instructional graphic design generation as multimodal layer generation in an editable format. DreamPoster \cite{hu2025dreamposter} studies image-conditioned generative poster design. These methods improve controllability and quality, but mainly focus on overall design synthesis rather than harmonizing potentially mismatched user-provided elements.

\noindent \textbf{Components-to-Design Generation:}
LaDeCo \cite{lin2024elementsdesignlayeredapproach} introduces a layered framework for automatic graphic design composition from multimodal elements. CreatiDesign \cite{zhang2025creatidesign} models this as a diffusion process jointly conditioned on input images, text, and layout constraints. FlexDM \cite{inoue2023flexiblemultimodaldocumentmodels} performs design synthesis by simultaneously masking and predicting element positions and text attributes. However, all of these methods preserve input components exactly or adjust only limited attributes such as typography and position, restricting their ability to handle visually mismatched inputs from diverse sources. Our work addresses this gap: unlike prior works that treat inputs as fixed, we adapt their appearance while preserving their identity, complementing existing layout and typography methods within end-to-end pipelines.

\section{Method}
\label{sec:method}


\begin{figure*}[t]
    \centering
    \includegraphics[width=0.85\textwidth, height=0.3\textheight, keepaspectratio]{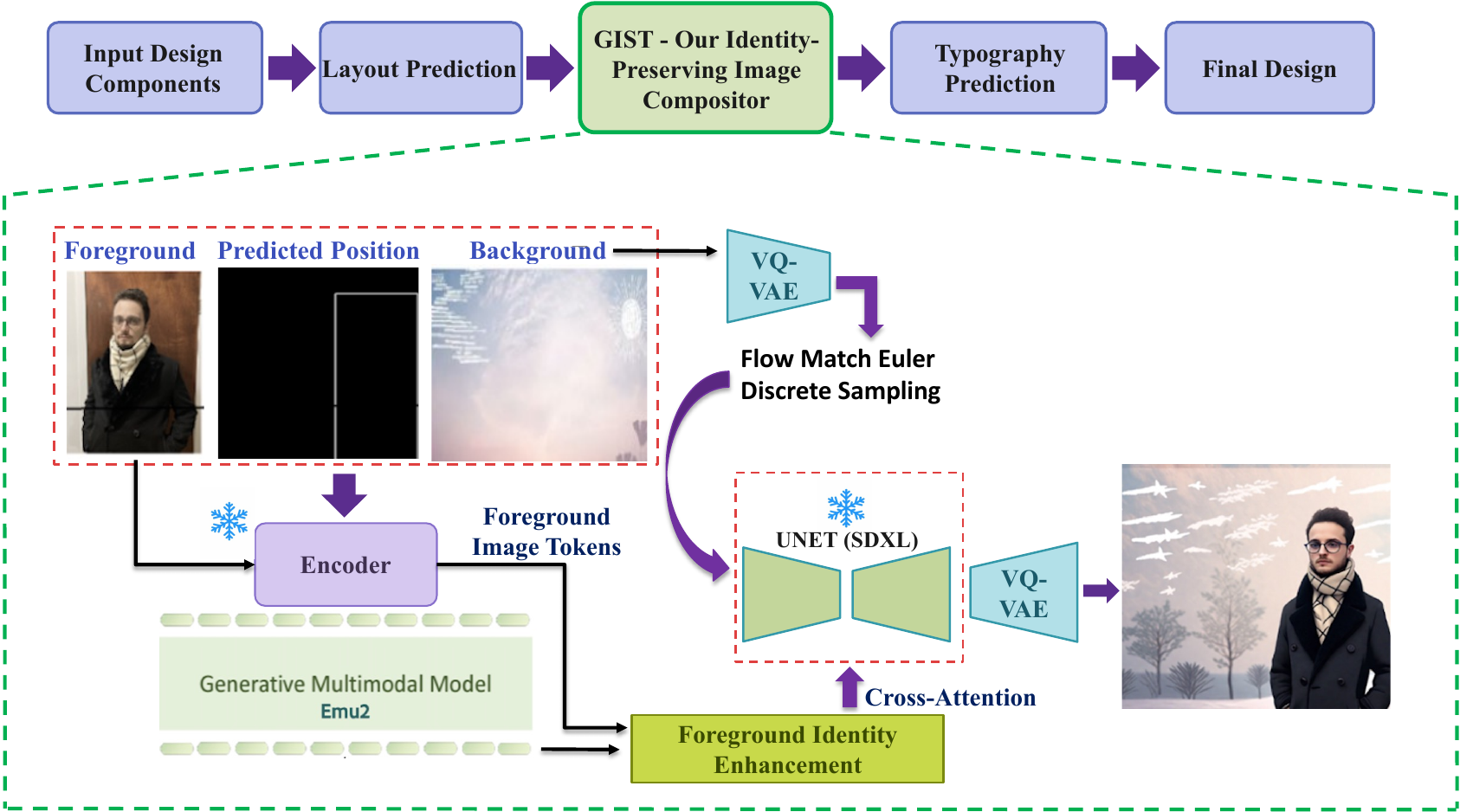} 
    \caption{%
\textbf{Overview of the Components-to-Design pipeline with \methodname{}.}
\methodname{} sits between layout prediction and typography as a plug-and-play compositing stage. Given a foreground element and its predicted position, the background canvas is inverted via Flow Matched Euler Discrete Sampling to initialize the denoising latent. In parallel, Emu-2's frozen visual encoder produces identity tokens from the foreground, corrected via our CA-guided token injection before conditioning SDXL's UNet alongside the generative tokens from Emu-2's LLaMA decoder. This process repeats sequentially for each element; the final composed backing image is passed to typography prediction to yield the complete design.}
    \label{fig:full_pipeline}
    \vspace{-12pt}
\end{figure*}

 				




    


As discussed in~\cref{sec:related}, existing components-to-design methods either preserve input assets exactly~\cite{lin2024elementsdesignlayeredapproach, inoue2023flexiblemultimodaldocumentmodels}, limiting their ability to handle visually mismatched inputs or generate assets from scratch~\cite{jia2023cole, inoue2024opencole}, sacrificing user-specified content. Neither paradigm addresses what we argue is a critical missing ingredient: \emph{stylizing and compositing user-provided image and SVG elements for visual harmony while preserving their semantic identity}.

To address this, we propose \textbf{\methodname{}}, a training-free, identity-preserving image  composition method (\cref{sec:identity}) designed to be \emph{plug-and-play}. Layout prediction, spatially arranging input visual elements has been well-explored by works such as LaDeCo~\cite{lin2024elementsdesignlayeredapproach} and Design-o-meter~\cite{goyal2024design}, and typography prediction, determining  placement, font, and color for text elements has been tackled by methods like  COLE~\cite{jia2023cole}, OpenCOLE~\cite{inoue2024opencole} and FlexDM~\cite{inoue2023flexiblemultimodaldocumentmodels}. \methodname{} sits between these two stages: given predicted elements and element positions as input, it produces a harmonized backing image ready for typography. Because it makes no assumptions about the surrounding modules, it can complement any existing components-to-design pipeline. We demonstrate  this in~\cref{sec:pipeline} by integrating \methodname{} into two distinct end-to-end pipelines built around LaDeCo and Design-o-meter.

\subsection{Identity-Preserving Composition}
\label{sec:identity}


Image composition involves placing a foreground element onto a background such that the result appears as a unified, coherent image. The challenge is threefold: (i)~the foreground's semantic identity must be preserved, (ii)~the background must remain faithful to the original, and (iii)~the composed result must be stylistically cohesive. Naive copy-paste satisfies (i) and (ii) but fails at (iii), producing jarring visual discontinuities when elements originate from different sources. Purely generative approaches achieve (iii) but often degrade (i) and (ii). Where the composition operation fits within the full design pipeline is detailed in~\cref{sec:pipeline}; here, we focus on the composition itself.


MLLMs are well-suited for this task, as their pretraining instills a strong prior over style, color, and texture coherence while supporting spatially grounded  generation. Visual elements (Images/SVGs) are composed sequentially given their predicted layouts, with the updated canvas serving as the background at each step; the full pipeline is detailed in~\cref{sec:pipeline}. At each step, rather than naively pasting, we prompt an MLLM with the background render, a foreground render and its caption (obtainable via any captioning model, like Qwen~\cite{bai2023qwen}), and the target position on the canvas, producing a harmonized composite. For this, we require a model whose internals are accessible enough to admit training-free manipulation, newer models such as FLUX Kontext~\cite{labs2025fluxkontext} produce 
good results but with limited training-free manipulations to make it stronger, making principled identity-preserving interventions difficult without fine-tuning. Emu-2~\cite{sun2024generative} uniquely exposes a 64-token bottleneck between its LLaMA decoder and SDXL renderer, which is precisely where we intervene since despite its stylization quality, off-the-shelf Emu-2 suffers from foreground and background identity loss. We address this with two training-free enhancements: \textbf{Cross-Attention Guided Token Injection}~\cite{hertz2022prompt, tumanyan2023plug, liu2024towards} (\cref{sec:ca_injection}) and \textbf{Latent Initialization}~\cite{mokady2023null, wallace2023edict, ju2023pnp} (\cref{sec:latent_init}).


\subsubsection{Cross-Attention Guided Token Injection}
\label{sec:ca_injection}

Off-the-shelf Emu-2 produces stylistically harmonized composites but suffers from foreground and background identity loss: the generated output may capture the right \emph{style} while failing to faithfully reproduce the foreground subject or the background canvas. The root cause is that the 64 image tokens produced by Emu-2's LLaMA decoder, which condition SDXL's rendering, are optimized for coherent generation rather than identity fidelity. We seek to correct this by injecting identity-specific information back into these tokens, but only where it matters spatially, so as not to destroy the model's stylization intent.

\noindent\textbf{Autoencoded Tokens as identity embeddings.}
A key property of Emu-2 is that its Visual Encoder (EVA-CLIP) and SDXL Decoder are jointly trained as an autoencoder: encoding an image directly through the visual encoder, bypassing LLaMA entirely, yields 64 tokens from which SDXL reconstructs the image near-perfectly. This means these tokens carry rich, fine-grained identity information. We exploit this: for any input image, encoding it through the visual encoder gives us an \emph{identity token set} that faithfully represents that image's appearance.

Concretely, we maintain two token sets: $T_\text{gen}$, produced by LLaMA model of Emu-2, from the full compositional prompt (carrying stylization intent), and $T_\text{auto}$, produced by encoding a naive foreground-on-background composite through the visual encoder alone (carrying identity). The challenge is deciding \emph{which} tokens in $T_\text{gen}$ to correct with identity from $T_\text{auto}$, and \emph{how much}, a global replacement would collapse stylization; no replacement leaves identity unprotected.

\noindent\textbf{Spatial relevance via cross-attention.}
The SDXL UNet's cross-attention maps provide exactly the signal we need. Each of the 64 tokens has a spatial CA map of shape $[H_\text{lat} \times W_\text{lat}]$, encoding how strongly each pixel location attends to that token during decoding. A token with high attention within the foreground bounding box is primarily responsible for rendering the foreground; one with high attention in the background region governs background appearance. We use these maps to assign each token a foreground and background relevance score:
\begin{align}
    r_\text{fg}[i] &= \frac{\max(\text{CA}[i] \odot \mathbf{m}_\text{fg})}{\max(\text{CA}[i])}, \\
    r_\text{bg}[i] &= \frac{\max(\text{CA}[i] \odot \mathbf{m}_\text{bg})}{\max(\text{CA}[i])},
\end{align}
where $\mathbf{m}_\text{fg}$ and $\mathbf{m}_\text{bg}$ are foreground and background binary masks. $\mathbf{m}_\text{fg}$ is derived from the element's alpha channel, or a hard bounding box mask if unavailable. To obtain these maps, we run a single lightweight scoring forward pass through the UNet at a mid-noise timestep using $T_\text{auto}$ as conditioning, and average CA maps across all attention layers.

\noindent\textbf{Selective identity injection.}
Armed with per-token relevance scores, we select the top-$N_\text{fg}$ tokens by $r_\text{fg}$ (set $\mathcal{S}_\text{fg}$) as the primary drivers of foreground appearance, and the top-$N_\text{bg}$ by $r_\text{bg}$ (set $\mathcal{S}_\text{bg}$) for background. We then blend the corresponding identity tokens from $T_\text{auto}$ into $T_\text{gen}$:
\begin{align}
    T_\text{final}[\mathcal{S}_\text{fg}] &= (1 {-} \beta_\text{fg}) \cdot T_\text{gen}[\mathcal{S}_\text{fg}] + \beta_\text{fg} \cdot T_\text{auto}[\mathcal{S}_\text{fg}], \\
    T_\text{final}[\mathcal{S}_\text{bg}] &= (1 {-} \beta_\text{bg}) \cdot T_\text{gen}[\mathcal{S}_\text{bg}] + \beta_\text{bg} \cdot T_\text{auto}[\mathcal{S}_\text{bg}],
\end{align}
with $\beta_\text{fg}{=}0.3$ and $\beta_\text{bg}{=}0.2$. All remaining tokens retain $T_\text{gen}$ unchanged. This injects identity precisely where each region is most represented in generation, without globally degrading stylization. The full procedure is summarized in~\cref{alg:token_injection}.

\begin{algorithm}[t]
\small
\caption{Cross-Attention Guided Token Injection}
\label{alg:token_injection}
\begin{algorithmic}[1]
\Require Foreground image, background canvas, target bounding box, masks $\mathbf{m}_\text{fg}$, $\mathbf{m}_\text{bg}$
\Ensure Blended token set $T_\text{final}$
\State $T_\text{gen} \gets$ LLaMA forward pass on compositional prompt \Comment{Stylization tokens}
\State $T_\text{auto} \gets$ Visual Encoder on fg-on-bg composite \Comment{Identity tokens}
\State Run UNet scoring pass with $T_\text{auto}$; collect and average CA maps across layers
\For{each token $i \in \{1,\ldots,64\}$}
    \State $r_\text{fg}[i] \gets \max(\text{CA}[i] \odot \mathbf{m}_\text{fg}) \;/\; \max(\text{CA}[i])$
    \State $r_\text{bg}[i] \gets \max(\text{CA}[i] \odot \mathbf{m}_\text{bg}) \;/\; \max(\text{CA}[i])$
\EndFor
\State $\mathcal{S}_\text{fg} \gets \operatorname{top\text{-}}N_\text{fg}$ indices by $r_\text{fg}$; \quad $\mathcal{S}_\text{bg} \gets \operatorname{top\text{-}}N_\text{bg}$ indices by $r_\text{bg}$
\State $T_\text{final} \gets T_\text{gen}$
\State $T_\text{final}[\mathcal{S}_\text{fg}] \gets (1{-}\beta_\text{fg})\,T_\text{gen}[\mathcal{S}_\text{fg}] + \beta_\text{fg}\,T_\text{auto}[\mathcal{S}_\text{fg}]$
\State $T_\text{final}[\mathcal{S}_\text{bg}] \gets (1{-}\beta_\text{bg})\,T_\text{gen}[\mathcal{S}_\text{bg}] + \beta_\text{bg}\,T_\text{auto}[\mathcal{S}_\text{bg}]$
\State \Return $T_\text{final}$
\end{algorithmic}
\end{algorithm}

\subsubsection{Background Fidelity: Latent Initialization}
\label{sec:latent_init}
To further improve background fidelity, we initialize the SDXL denoising process from a partially noised version of the background canvas rather than pure noise. Specifically, we encode the background through the VQ-VAE encoder and apply the Flow Matched Euler Discrete Scheduler~\cite{esser2024scaling} (the same scheduler used in Emu-2's training) to invert it. This linear interpolation between data and noise preserves more structural information than alternatives such as DDIM~\cite{song2020denoising} or LCM~\cite{luo2023latent} inversion (see Supp.\ for comparisons). Starting denoising from this initialized latent biases generation toward the original background appearance while still allowing the model to harmonize foreground and background.

\subsection{Augmenting Design Generation with GIST}
\label{sec:pipeline}
The composition method described above operates on individual image elements given a background and target position. To produce complete graphic designs from multimodal components, current Components-to-Design frameworks~\cite{lin2024elementsdesignlayeredapproach} do layout prediction of all elements (Images, SVG's) and Typography prediction (Positions, Font and Colours of texts) and render. We embed \textbf{image composition} in between layout prediction and \textbf{typography prediction} (see~\cref{fig:full_pipeline}). Because our composition module only requires element positions as input and produces a composed backing image as output, it slots into any existing pipeline without modification to the surrounding stages. We demonstrate this with two configurations representing different design paradigms.

\subsubsection{Layout Prediction}
\label{sec:layout}
We integrate with two layout methods. \textbf{LaDeCo}~\cite{lin2024elementsdesignlayeredapproach} is a state-of-the-art LMM-based method that predicts element positions, sizes, and layer ordering from input components, and additionally provides typography attributes. \textbf{Design-o-meter}~\cite{goyal2024design} takes a different approach: it trains a Siamese aesthetic scorer on good/bad design pairs and uses NSGA-II~\cite{deb2000fast} optimization to refine layout coordinates by maximizing the learned aesthetic objective. Both output spatial coordinates for each visual element, which serve as direct input to our composition module.

\subsubsection{Compositing}
\label{sec:compositing}



Using the layout coordinates from~\cref{sec:layout}, we apply the method from~\cref{sec:identity} to sequentially compose each visual element onto the canvas. We initialize the canvas with the root background element. For each subsequent element in the predicted layer order, we: (1)~caption the foreground element, (2)~compute $T_\text{auto}$ and $T_\text{gen}$, (3)~perform token injection and SA manipulation, (4)~invert the current canvas to obtain the initial latent, and (5)~generate the composed image with SDXL. For SVG elements, we remove the background from the generated output before pasting it back at the target position; for image elements, the latent initialization ensures natural blending. The canvas is updated after each element, so later elements are conditioned on all previously composed ones.

\subsubsection{Typography}
\label{sec:typography}


Given the composed backing image, the final stage predicts text placement, font, and color for each input text element. Since diffusion models still struggle with artifact-free stylized text rendering, we rely on dedicated typography modules followed by a graphic renderer. For the LaDeCo-based pipeline, we use LaDeCo's own typography predictions. For the Design-o-meter pipeline, we train a typography LMM inspired by COLE~\cite{jia2023cole}: we finetune InternLM-XComposer2~\cite{dong2024internlm} with a two-stage strategy, first grounding the model on single-element typography detection, then finetuning on full-design typography prediction (details in Supp.). The plug-and-play nature of our composition module means it is agnostic to the typography method used downstream.

\section{Experiments and Results}
\label{sec:exp}

We evaluate our method along two axes: (i)~the quality of end-to-end designs produced by the full pipelines when \methodname{} is plugged into them, and (ii)~the effectiveness of our identity-preserving composition module in isolation. We first describe the experimental setup, then present quantitative and qualitative results for complete design generation, followed by a focused analysis of the composition module and ablations validating our design choices.

\subsection{Experimental Setup}
\label{sec:setup}


We evaluate on the Crello test set~\cite{yamaguchi2021canvasvae}, comprising 1{,}500 graphic designs with layer-wise metadata, providing background images, foreground image elements, SVG shapes, and text assets extracted from real designs. Since these components originate from the same ground-truth design, they are already stylistically compatible, making our evaluation \emph{conservative}: our method targets the harder real-world setting where inputs come from disparate sources.

\noindent\textbf{Baselines.}
We compare against the following methods:
\textbf{FlexDM}~\cite{inoue2023flexiblemultimodaldocumentmodels}: a multimodal transformer that jointly predicts layout and typography through masked prediction.
\textbf{GPT-4o}: layout planning via GPT-4o with few-shot prompting, followed by direct alpha-composite paste of elements.
\textbf{LaDeCo}~\cite{lin2024elementsdesignlayeredapproach}: state-of-the-art LMM-based layout and typography prediction, with elements naively pasted at predicted positions. 
\textbf{Design-o-meter}~\cite{goyal2024design}: layout-only optimization via NSGA-II, no typography.
\textbf{GT}: ground-truth Crello designs as a human-quality reference.

\begin{figure}
    \centering
 \includegraphics[width=0.48\textwidth]{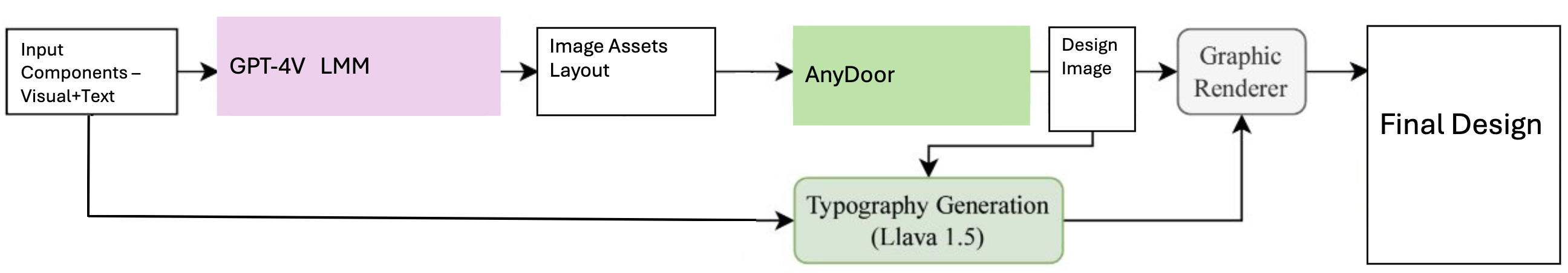}
    \caption{OpenCOLE++ pipeline overview} 
    \label{fig:opencole++}
\end{figure}

Since no prior work directly addresses components-to-design generation \textbf{with composition}, we evaluate across two distinct comparison axes. The first is an \emph{isolated} comparison against \textbf{LaDeCo}: we share the same layout and typography backbone and swap only the compositing step, directly measuring the contribution of \methodname{} over naive pasting. This is intentional, our goal 
is not to compete with layout methods but to complement them, and this setup isolates exactly that effect. The second is a \emph{holistic} comparison against \textbf{OpenCOLE++}, a different baseline we construct by adapting OpenCOLE~\cite{inoue2024opencole} to accept input components (see~\cref{fig:opencole++}): GPT-4V replaces OpenCOLE's design planner for layout prediction, and AnyDoor~\cite{chen2024anydoor} is its image generator for diffusion-based composition (We intentionally use AnyDoor instead of naive pasting, to compete with \methodname{}). We pair this with our Design-o-meter~\cite{goyal2024design}-based pipeline, which uses \methodname{} for composition and our trained typography module. This holistic comparison reflects the compounded effect of swapping every pipeline 
component simultaneously, and benchmarks our full system against an alternative paradigm for design composition.

\noindent\textbf{Evaluation metrics.}
Following prior work~\cite{jia2023cole, cheng2024graphic}, we employ LLaVA-OV~\cite{li2024llava} for automated design evaluation along two protocols:
(1)~\emph{Aspect-wise rating}: each design is scored on a 1-10 scale across five aspects - Design \& Layout, Content Relevance, Typography \& Color, Graphics \& Imagery, and Innovation \& Originality.
We don't report the Design \& Layout, Typography \& Color numbers as we leverage the layout \& typography models of the baseline, in this case, LaDeCo, so the metric is unchanged. 
(2)~\emph{Pairwise preference}: LLaVA-OV is asked to choose the more aesthetic design between our output and the LaDeCo paste-only baseline, over 1{,}500 paired comparisons.
For the OpenCOLE++ comparison, we use GPT-4V~\cite{achiam2023gpt} for both rating and pairwise voting, following the evaluation protocol of COLE~\cite{jia2023cole}.

\subsection{Quantitative Evaluation}
\label{sec:main_results}

\subsubsection{Comparison with LaDeCo}

\begin{table}[t]
\centering
\small
\caption{LLaVA-OV ratings (1--10) on the Crello test set. We report Content Relevance, Graphics \& Imagery, and Innovation \& Originality - the aspects most sensitive to our compositing contribution, along with the mean across all five aspects. Design \& Layout and Typography \& Color are omitted from individual columns as they primarily reflect the shared LaDeCo layout backbone; they are included in the mean.}
\label{tab:main_results}
\setlength{\tabcolsep}{3.9pt}
\begin{tabular}{lcccc}
\toprule
Method & Content & Graphics & Innovation & Mean$\uparrow$ \\
\midrule
FlexDM~\cite{inoue2023flexiblemultimodaldocumentmodels} & 5.29 & 5.09 & 4.54 & 5.13 \\
GPT-4o & 6.49 & 6.27 & 5.69 & 6.32 \\
LaDeCo ~\cite{lin2024elementsdesignlayeredapproach} & \textbf{7.96} & 7.74 & 6.93 & 7.77 \\
\textbf{Ours (w/ LaDeCo)} & 7.89 & \textbf{7.83} & \textbf{7.05} & \textbf{7.79} \\
\midrule
GT & 8.17 & 7.93 & 7.15 & 7.95 \\
\bottomrule
\end{tabular}
\vspace{-8pt}
\end{table}

\begin{figure*}[h]
    \centering
    \includegraphics[width=\textwidth, height=0.90\textheight, keepaspectratio]{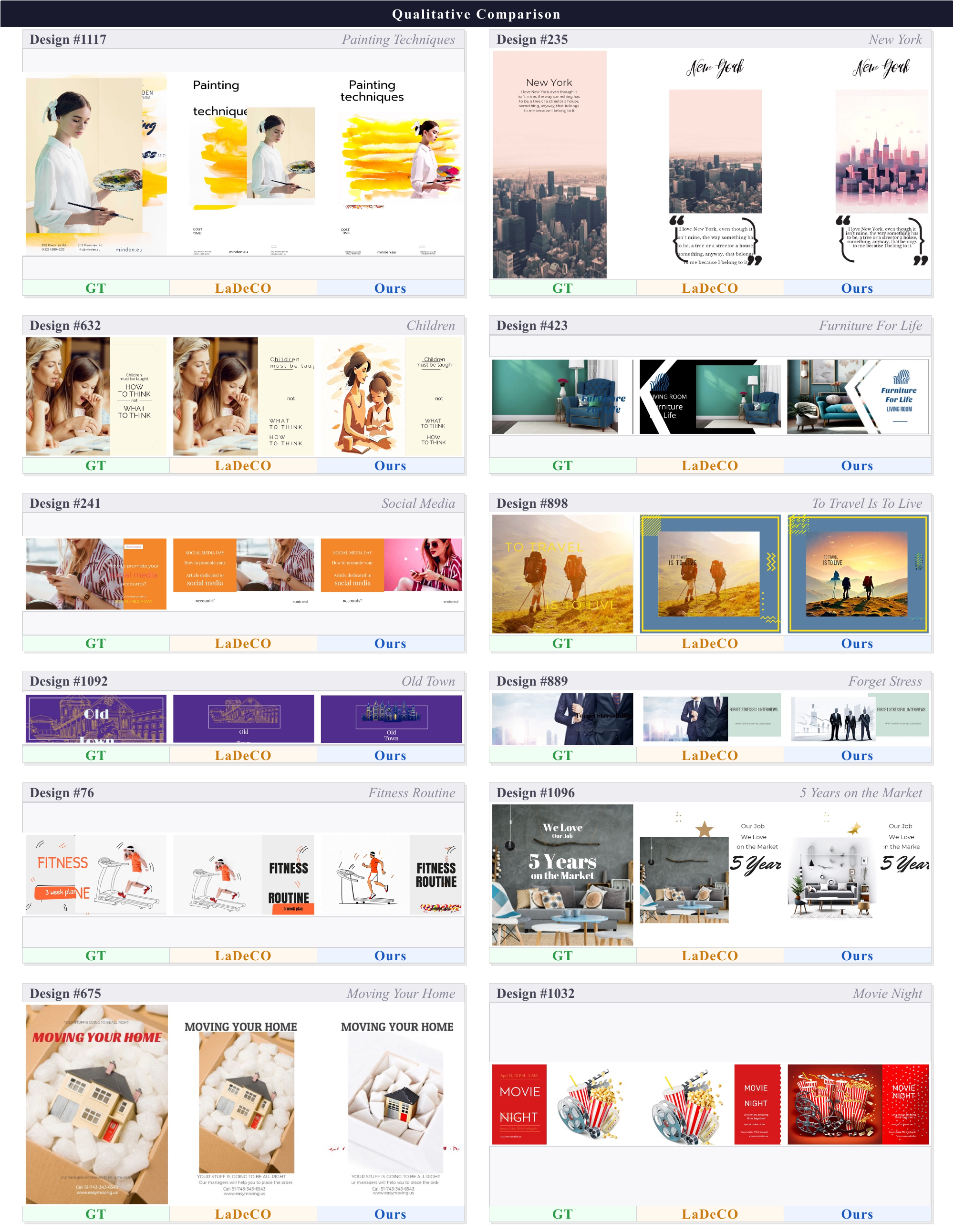}
    \caption{Qualitative comparison on Crello test samples. \textbf{Columns}: Input components, LaDeCo (paste-only), Ours (LaDeCo w/ \methodname{}, and Ground Truth. In paste-only results, elements appear as flat cut-outs with visible edge discontinuities and color mismatches. Our method harmonizes lighting, color palette, and texture across elements, producing compositions that are visually cohesive and closer to GT. }
    \label{fig:qualitative_main}
    \vspace{-6pt}
\end{figure*}

\begin{figure}[h]
    \centering
 \includegraphics[width=0.48\textwidth]{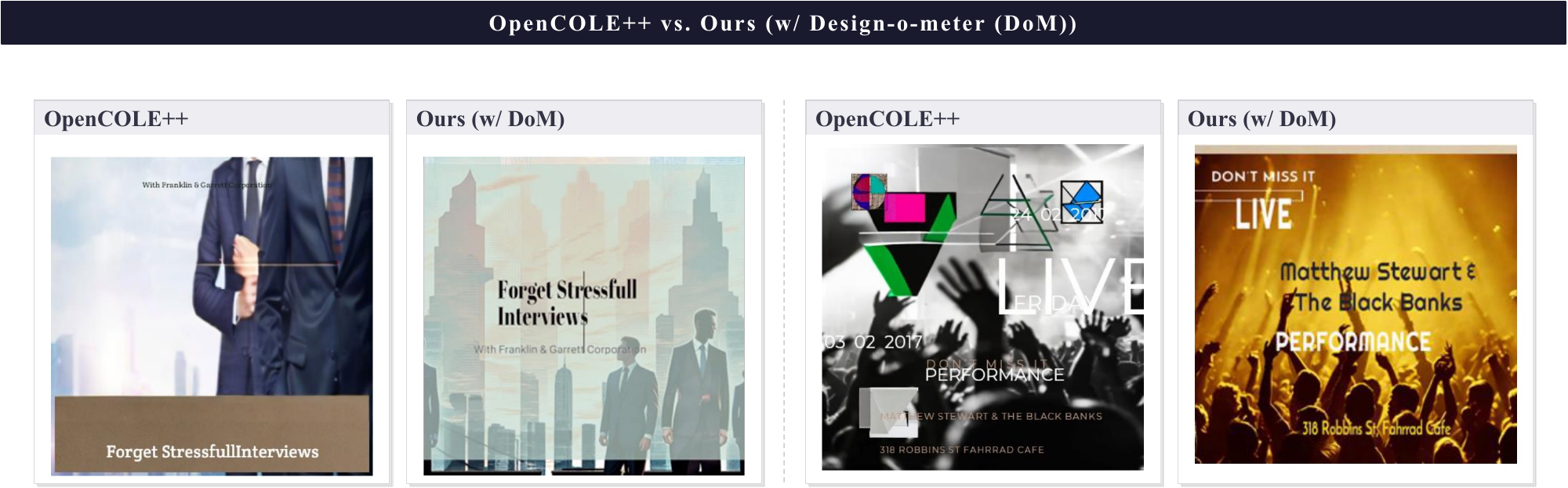}
 \vspace{-15pt}
    \caption{OpenCOLE++ vs Ours (w/ Design-o-meter)} 
    \label{fig:opencole++quali}
    \vspace{-3pt}
\end{figure}

\begin{figure}[h]
    \centering
 \includegraphics[width=0.48\textwidth]{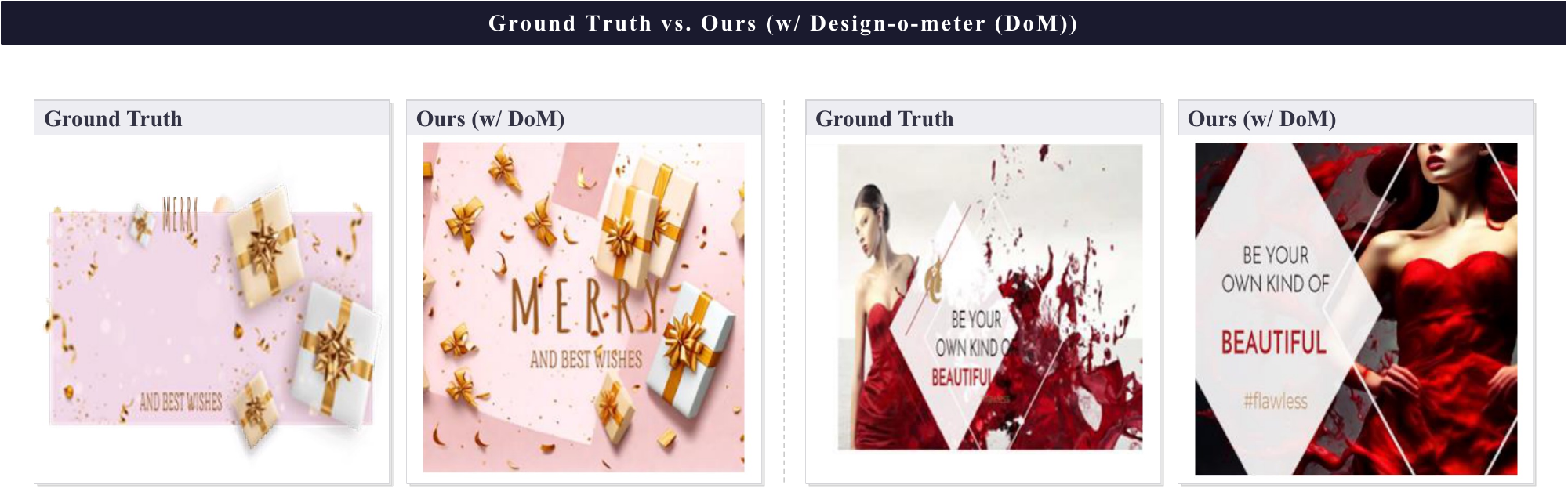}
 \vspace{-15pt}
    \caption{Ground Truth designs vs Ours (w/ Design-o-meter)} 
    \label{fig:gtquali}
    \vspace{-3pt}
\end{figure}

\noindent\textbf{Quantitative analysis.}
\cref{tab:main_results} summarizes the LLaVA-OV ratings. Our method (LaDeCo with \methodname{}) outperforms vanilla LaDeCo on \emph{Graphics \& Imagery} (7.83 vs.\ 7.74) and \emph{Innovation \& Originality} (7.05 vs.\ 6.93), the two dimensions most directly affected by image-level compositing, while remaining within noise on Content Relevance (7.89 vs.\ 7.96). The overall mean is essentially tied (7.79 vs.\ 7.77). This is the intended behavior: our compositor enhances aesthetic harmony and visual richness \emph{without degrading content fidelity}. The gap to GT is narrow, 0.10 on Graphics \& Imagery and 0.16 on the overall mean, indicating that our pipeline produces near-professional-quality designs. FlexDM and GPT-4o are substantially weaker across all dimensions, confirming that the LaDeCo layout backbone is important and that our gains are attributable to the compositing stage.

\noindent\textbf{Design Improvement.}
Out of 1{,}500 designs in the Crello test set, \methodname{} meaningfully improves upon the LaDeCo naive-pasting baseline in \textbf{40.3\%} of cases, with an additional 14.7\% 

\subsubsection{Comparison with OpenCOLE++}

\begin{table}[t]
\centering
\small
\caption{GPT-4V ratings (1--10) comparing our Design-o-meter (DoM) based pipeline against OpenCOLE++. Aspects: (i)~Design \& Layout, (ii)~Content Relevance, (iii)~Typography \& Color, (iv)~Graphics \& Imagery, (v)~Innovation \& Originality.}
\label{tab:opencole}
\resizebox{\columnwidth}{!}{%
\begin{tabular}{lcccccc}
\toprule
Method & (i) & (ii) & (iii) & (iv) & (v) & Mean$\uparrow$ \\
\midrule
OpenCOLE++ & 4.6 & 5.2 & 4.0 & 5.2 & 5.3 & 4.9 \\
\textbf{Ours (w/ DoM)} & \textbf{5.8} & \textbf{6.2} & \textbf{5.1} & \textbf{6.7} & \textbf{5.8} & \textbf{5.9} \\
\bottomrule
\end{tabular}}
\vspace{-6pt}
\end{table}

Quantitatively (\cref{tab:opencole}), Design-o-meter+\methodname{}+our typography model outperforms OpenCOLE++ across all five aspects, with particularly large gains on Graphics \& Imagery (+1.5) and Content Relevance (+1.0), and achieves a 1.0-point higher mean rating (5.9 vs.\ 4.9). In GPT-4V pairwise voting, our outputs are preferred \textbf{71.43\%} of the time. Notably, this comparison uses our Design-o-meter-based pipeline rather than the LaDeCo backbone of~\cref{tab:main_results}, demonstrating that \methodname{} produces strong results regardless of the layout module, a direct validation of its plug-and-play nature.

\subsection{Qualitative Results}
\label{sec:qualitative}


\cref{fig:qualitative_main} shows representative examples comparing our method with LaDeCo against vanilla LaDeCo (naive paste-only) and GT. In the paste-only outputs, elements appear as flat overlays: edges are visible, lighting is inconsistent across layers, and color palettes clash. Our compositor harmonizes these elements into a unified scene, lighting becomes consistent, colors blend naturally, and the overall composition feels integrated rather than assembled. Compared to GT, our method occasionally produces more stylized variants; this is a natural consequence of the generative process, and accounts for some of the pairwise losses on Crello's pre-matched components. 

\noindent\textbf{Comparison with OpenCOLE++.}
~\cref{fig:opencole++quali}, ~\cref{fig:gtquali},  compares our method against OpenCOLE++ and Ground Truth Crello Designs. While AnyDoor~\cite{chen2024anydoor} (not all examples shown in the image) produces stylistically coherent outputs, it frequently fails to preserve the identity of foreground elements. Objects are altered, faces lose likeness, and branded assets become unrecognizable. Our method, through CA-guided token injection, achieves harmonization \emph{without} sacrificing identity.


\subsection{Ablation Studies}
\label{sec:ablations}

We ablate the two key components of our identity-preserving composition module: Cross-Attention Guided Token Injection (\cref{sec:ca_injection}) and Latent Initialization (\cref{sec:latent_init}).

\noindent\textbf{Identity preservation.}
To isolate the composition module from the full pipeline, we evaluate foreground identity retention on two controlled tasks: \emph{Human Face Generation (HFG)}, compositing face images onto diverse backgrounds, and \emph{Specific Object Generation (SOG)}, compositing identifiable everyday objects (data from CelebA~\cite{liu2015deeplearningfaceattributes}, Stanford Background~\cite{5459211}, and DreamEditBench~\cite{li2023dreameditsubjectdrivenimageediting}; details in Supp.). We measure cosine similarity, Manhattan distance, and Euclidean distance between embeddings of the input foreground and the composed output.

\begin{table}[t]
\centering
\small
\caption{Identity preservation metrics for our compositor vs.\ off-the-shelf Emu-2. Cosine similarity ($\uparrow$) is computed on face embeddings for HFG; Manhattan ($\downarrow$) and Euclidean ($\downarrow$) distances are on image embeddings for HFG and SOG respectively.}
\label{tab:identity}
\setlength{\tabcolsep}{7pt}
\begin{tabular}{lccc}
\toprule
Method & Cos.\ Sim.\ $\uparrow$ & Manhattan $\downarrow$ & Euclidean $\downarrow$ \\
\midrule
Emu-2~\cite{sun2024generative} & 0.09 & 554.52 & 961.07 \\
\textbf{\methodname{} (Ours)} & \textbf{0.13} & \textbf{540.45} & \textbf{943.91} \\
\bottomrule
\end{tabular}
\vspace{-4pt}
\end{table}

\cref{tab:identity} shows that our method consistently outperforms off-the-shelf Emu-2 across all metrics, achieving higher cosine similarity and lower distances, confirming that our training-free interventions meaningfully improve foreground identity retention during composition.

\noindent\textbf{Component ablation.}
\cref{tab:ablation} isolates the contribution of each proposed component by measuring face embedding cosine similarity on the HFG task. Removing either Cross-Attention Guided Token Injection 
causes a substantial drop in identity preservation, with both components contributing roughly equally. The full method achieves the highest similarity, validating that both interventions are necessary and complementary.

\begin{table}[t]
\centering
\small
\caption{Ablation of composition components on HFG cosine similarity ($\uparrow$).}
\label{tab:ablation}
\setlength{\tabcolsep}{15pt}
\begin{tabular}{lc}
\toprule
Configuration & Cos.\ Sim.\ $\uparrow$ \\
\midrule
\textbf{Full method (\methodname{})} & \textbf{0.13} \\
\quad w/o CA-Guided Token Injection & 0.04 \\
\bottomrule
\end{tabular}
\vspace{-4pt}
\end{table}




\section{Conclusion and Limitations}
\label{sec:conclusion}

We presented \methodname{}, a training-free, identity-preserving image compositor that stylizes and harmonizes user-provided visual elements while retaining their semantic identity, and plugs seamlessly into any existing components-to-design pipeline. We hope this work motivates broader attention to input-conditioned composition as a critical missing ingredient in automated graphic design. Diffusion artifacts, unaddressed textual composition, and architectural ties to Emu-2's bottleneck remain open challenges; extending our approach to newer unified generative models such as FLUX~\cite{labs2025fluxkontext} and Janus-Pro~\cite{chen2025janus} is a natural next step.

{
    \small
    \bibliographystyle{ieeenat_fullname}
    \bibliography{main}
}


\end{document}

%% file: preamble.tex
\newcommand{\methodname}{\textbf{GIST}}

